\documentclass[conference, A4]{IEEEtran}
\IEEEoverridecommandlockouts
\usepackage{subfigure} 
\usepackage{cite}
\usepackage{amsmath,amssymb,amsfonts}
\usepackage{algorithmic}
\usepackage{graphicx}
\usepackage{textcomp}
\usepackage{xcolor}
\usepackage{mathrsfs}
\usepackage{footnote}
\usepackage{balance}
\usepackage{url}

\def\BibTeX{{\rm B\kern-.05em{\sc i\kern-.025em b}\kern-.08em
    T\kern-.1667em\lower.7ex\hbox{E}\kern-.125emX}}
\begin{document}

\title{\LARGE \bf PT-ResNet: Perspective Transformation-Based Residual \\Network for Semantic Road Image Segmentation}

\author{
	Rui Fan$^{1*}$, Yuan Wang$^{1*}$, Lei Qiao$^{2}$, Ruiwen Yao$^{2}$, Peng Han$^{2}$, Weidong Zhang$^{2}$, Ioannis Pitas$^{3}$, Ming Liu$^{1}$\\

$^{1}$Robotics Institute, the Hong Kong University of Science and Technology, Hong Kong.\\

$^{2}$Department of Automation, Shanghai Jiao Tong University, Shanghai 200240, China.\\

$^{3}$Department of Informatics, Aristotle University of Thessaloniki, Thessaloniki, Greece.
\\
% yaoruiwen88@foxmail.com,\\ han$_$ipac@sjtu.edu.cn, wdzhang@sjtu.edu.cn,
{eeruifan@ust.hk, ywangeq@connect.ust.hk, qiaolei2008114106@gmail.com, yaoruiwen88@foxmail.com,}\\ {han\_ipac@sjtu.edu.cn, wdzhang@sjtu.edu.cn, eelium@ust.hk, pitas@csd.auth.gr}\\ 

\thanks{*This two authors are joint first authors.}
\vspace{-1.5em}
}

\maketitle
\begin{abstract}
Semantic road region segmentation is a high-level task, which paves the way towards road scene understanding. This paper presents a residual network trained for  semantic road segmentation. Firstly, we represent the projections of road disparities in the v-disparity map  as a linear model, which can be estimated by optimizing the v-disparity map using dynamic programming. This linear model is then utilized to reduce the redundant information in the left and right road images. The right image is also transformed into the left perspective view, which greatly enhances the  road surface similarity between the two images.  Finally, the processed stereo images and their disparity maps are concatenated to create a set of 3D images, which are then utilized to train our neural network. The experimental results illustrate that our network achieves a maximum F1-measure of approximately $\boldsymbol{91.19\%}$, when analyzing the images from the KITTI road dataset. 
\end{abstract}
%\begin{IEEEkeywords}
%residual network, semantic road segmentation, v-disparity, dynamic programming.
%\end{IEEEkeywords}

\section{Introduction}
\label{sec.introduction}
Autonomous driving technology has been developing rapidly, since Google launched its self-driving car project in 2009 \cite{Brink2017}. In recent years,  industry titans, such as Waymo and Tesla, race to commercialize autonomous vehicles (AVs) \cite{Fan2019, Fan2018}. However, a number of high-profile experimental accidents that occurred in the last year and have called into question whether the autonomous driving technology is mature enough for employment \cite{lin2016ethics}. Therefore, most researchers believe that in  the next few years the research  on autonomous driving  should focus on developing advanced driver assistance systems (ADASs) \cite{David2019}, such as lane marking detection \cite{Ozgunalp2017}, road surface 3D reconstruction \cite{Fan_2019_CVPR_Workshops}, 2D/3D object detection \cite{du2018general}, etc.

Visual environment perception (VEP) is a key component of ADAS \cite{ Yan2018, fan2018real}. After learning from a large amount of labeled training data, VEP can extract useful road environment information, e.g., free space areas and pedestrians, from road images  \cite{Baek2018}, semantic image region segmentation can provide useful information by partitioning an image into semantically meaningful regions and classifying them into one of the pre-defined categories \cite{Long2015}. State-of-the-art semantic segmentation algorithms are generally based on fully convolutional networks (FCNs) \cite{Long2015}, which are an extension of convolutional neural network (CNN). FCNs utilize  classical CNNs to learn image feature representations, but the input images can be of any sizes. FCNs perform image upsampling to produce a probability mask with the same size as the input image \cite{Mendes2016}. 
%\vspace{-2em}

\begin{figure*}[!t]
	\centering
	\includegraphics[width=1\textwidth]{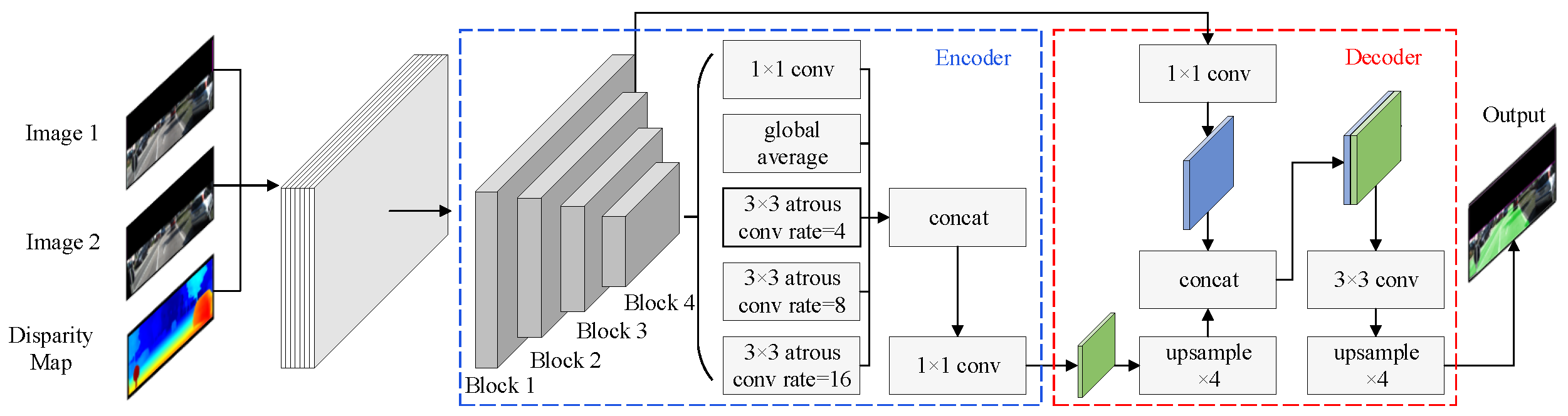}
	\caption{PT-ResNet structure.}
	\label{fig.structure}
\end{figure*}
\begin{figure}[!t]
	\centering
	\includegraphics[width=0.49\textwidth]{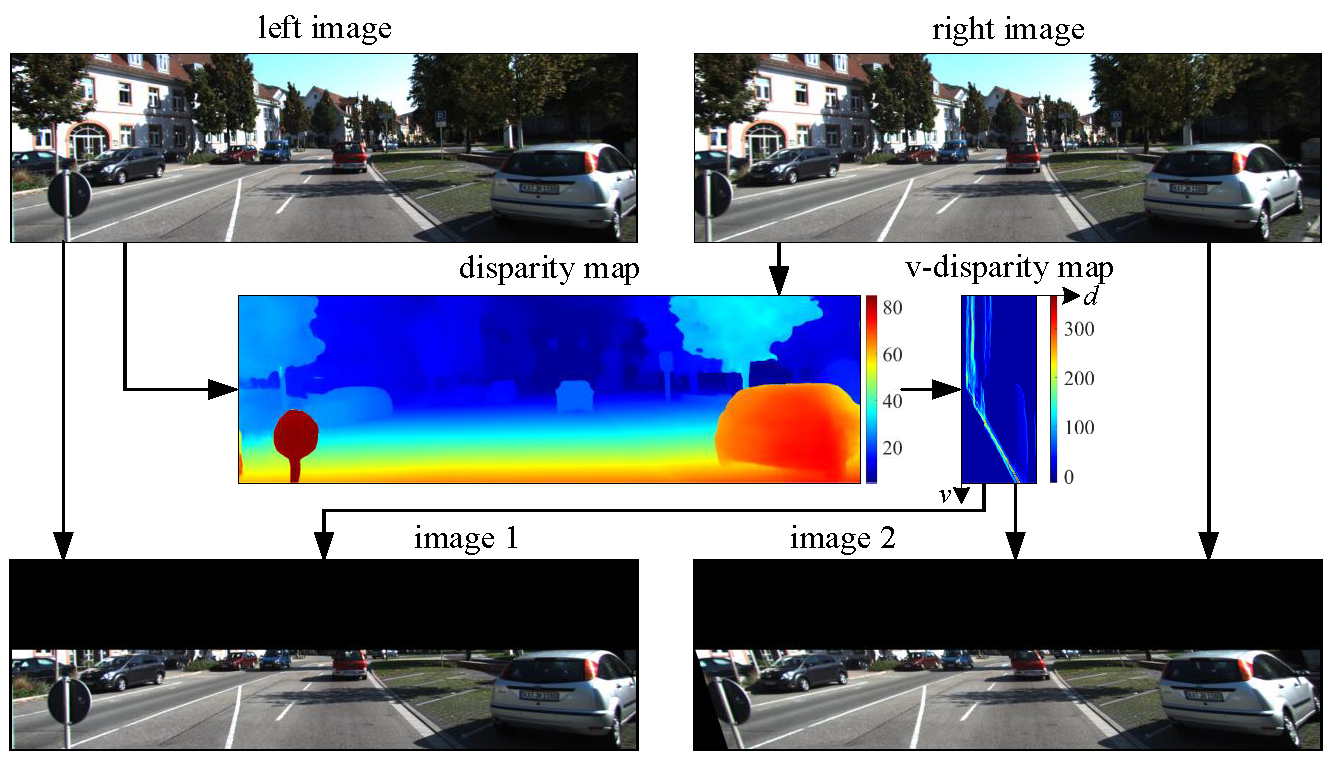}
	\caption{Training data pre-processing and generation.}
	\label{fig.data_generation}
\end{figure}

FCN-LC \cite{Mendes2016} is a classical FCN used for semantic road image segmentation. FCN-LC utilizes a network-in-network architecture \cite{Mendes2016} to learn road region segmentation from labeled training image data. This allows fast inference, even for large contextual image window sizes \cite{Mendes2016}. In addition, in recent years, a number of conditional random field (CRF)-based neural networks, e.g., PGM-ARS \cite{Passani2015}, Hybrid \cite{Xiao2018} and StixelNet \cite{Levi2015} have been proposed for semantic road image segmentation. PGM-ARS \cite{Passani2015} and StixelNet \cite{Levi2015} were trained using monocular images, while Hybrid \cite{Xiao2018} also employed the 3D road scene information acquired using LiDAR for training.  Furthermore,  stereo vision \cite{Wang2014, Vitor2014} was used to improve road segmentation performance. 
For example, a so-called BM neural network \cite{Wang2014}
selects a region of interest (ROI) in an image, by analyzing the v-disparity information. Such ROI information greatly minimizes the number of incorrectly segmented pixels. 
Furthermore, a so-called HistonBoost network  \cite{Vitor2014} post-processes such ROIs using watershed transformation and morphological filtering \cite{Pitas2000}. In this paper, we draw on the success of \cite{Wang2014, Vitor2014} and present a perspective transformation (PT)-based deep convolutional network for road semantic segmentation. It is designed using the residual network (ResNet) from DeepLab \cite{Chen2018}. 
The structure of our proposed  network is shown in Fig. \ref{fig.structure}.

The rest of the paper is organized as follows: Section \ref{sec.method} introduces the proposed PT-ResNet. In Section \ref{sec.experimental_results}, the experimental results are illustrated and the performance of the proposed approach is evaluated. Section \ref{sec.conclusion} contains conclusion and some recommendations for future work. 
\section{Methodology}
\label{sec.method}

\subsection{Training Data Pre-Processing and Generation}
\label{sec.training_data}
In this paper, the proposed semantic segmentation method focuses entirely on the road surface, which can be treated as a ground plane.  According to the perspective transformation algorithm presented in \cite{Fan2018a}, a right image can be transformed into its left view using the disparity projection model. This can greatly enhance the similarity of the road surface between the stereo images \cite{Fan2018a}.  Therefore, in this paper, we first utilize PSMNet \cite{Chang2018} to estimate dense disparity maps (see Fig. \ref{fig.data_generation}). A v-disparity map is then created by computing the histograms $\hat{p}(d,v)$ of each horizontal row $v$ of the disparity map \cite{fan2019road}. To find the path corresponding to the road disparity projection  in the v-disparity map, we utilize dynamic programming (DP) to search for every possible solution \cite{fan2019pothole}:
\begin{equation}
\begin{split}
E(d,v)=-\hat{p}(d,v)
+\min_{\tau=0}^{\tau_\text{max}}[E(d+1,v-\tau)-\lambda\tau],
\end{split}
\label{eq.DP}
\end{equation}
where $\hat{p}(d,v)$ represents the histogram value at $(d,v)$ in the v-disparity map, $\lambda$ is a smoothness term, $\tau_\text{max}$ is the maximum search range \cite{fan2018novel}. $E$ represents the energy of each possible solution. The path corresponding to the road disparity projection is generally represented using a linear polynomial \cite{Fan2018a}:
\begin{equation}
f(v)=\alpha_0+\alpha_1 v.
\label{eq.linear_model}
\end{equation}
The vertical coordinate of the vanishing point, i.e., $v_{py}$, can be estimated using (\ref{eq.linear_model}). As the vertical coordinates of the road pixels are always larger than $v_{py}$, the image region above the vanishing point can be  removed from the left and right images (see Fig. \ref{fig.data_generation}). Then, we utilize our previous algorithm \cite{Fan2018a} to transform the perspective view of the right image.  This algorithm  improves the  road surface similarity in the stereo images, but also distorts obstacles, such as vehicles and trees.
Finally, the processed stereo images and the left disparity maps are concatenated together to generate a set of 3D images with seven channels,  which are then utilized to train the neural network.  

\subsection{PT-ResNet Structure}
\label{sec.network}
\begin{figure}[t]
	\centering
	\includegraphics[width=0.49\textwidth]{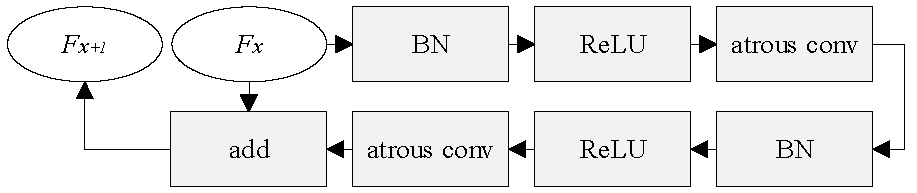}
	\caption{The structure of each block unit in Fig. \ref{fig.structure}.}
	\label{fig.block_unit}
\end{figure}
\begin{figure*}[t]
	\begin{center}
		\centering
		\includegraphics[width=0.98\textwidth]{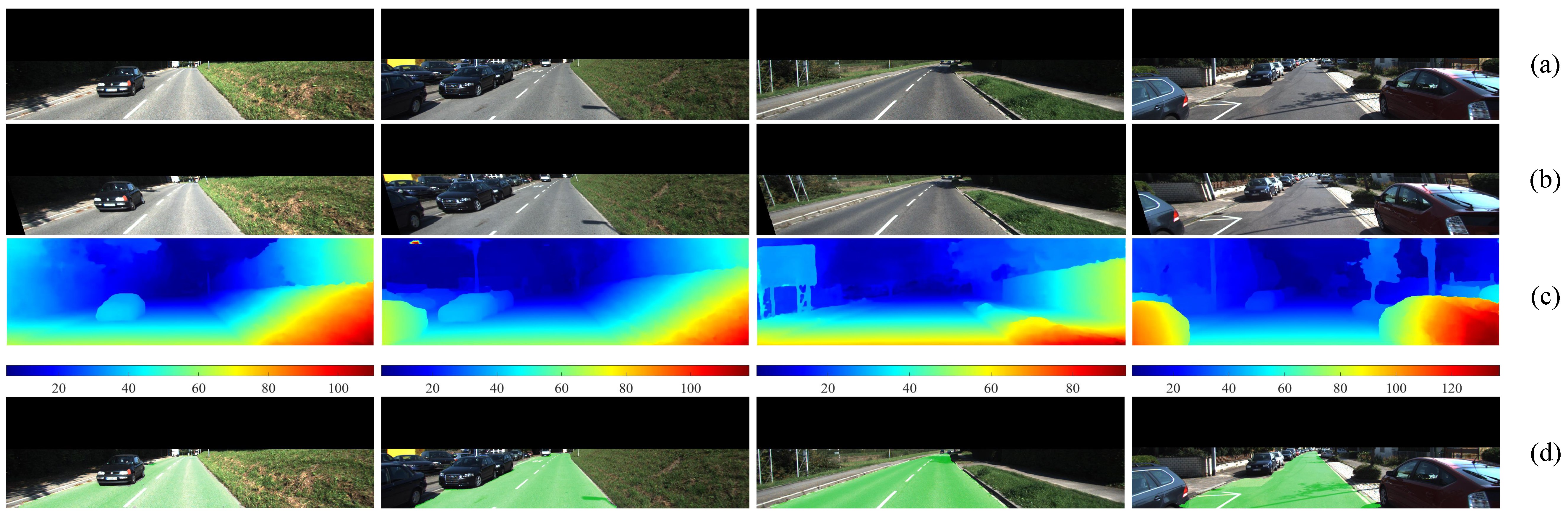}
		\centering
		\caption{Experimental results of road semantic segmentation (threshold is set to 0.9). The green areas in the fourth row are the segmented road surfaces. (a) Processed left images. (b) Transformed right images. (c) Disparity maps. (d) Segmentation results.  }
		\label{fig.experimental_results}
	\end{center}
\end{figure*}
\begin{figure*}[t]
	\centering
	\subfigure[]
	{\includegraphics[width=0.35\textwidth]{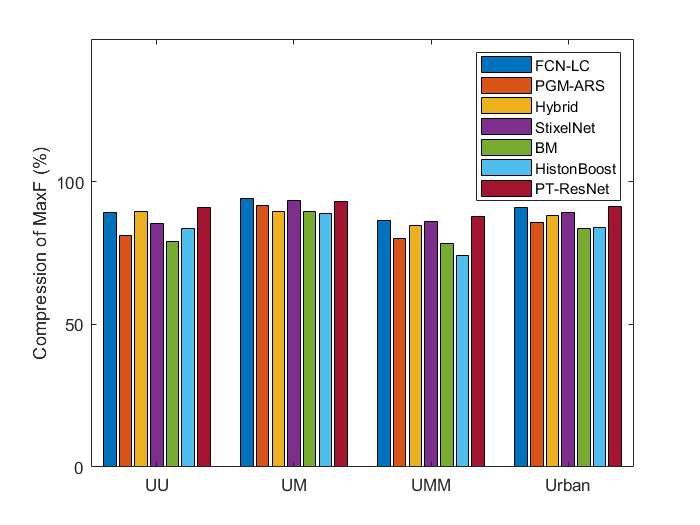}
		\label{fig.maxf}}
	\subfigure[]
	{\includegraphics[width=0.35\textwidth]{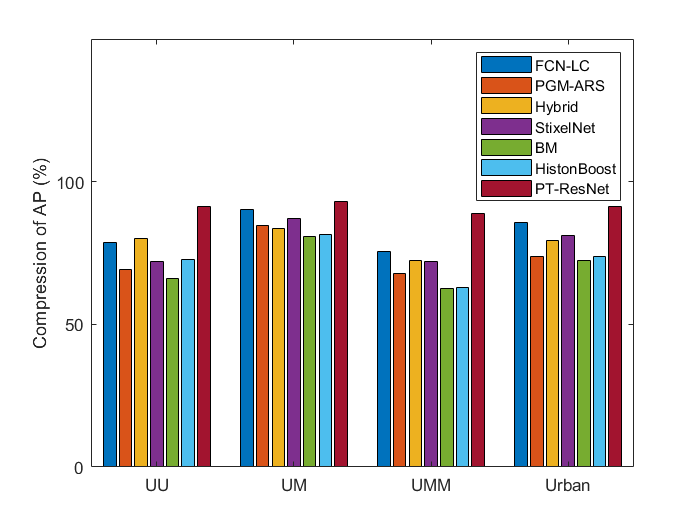}
		\label{fig.ap}}
	\\
	\subfigure[]
	{\includegraphics[width=0.35\textwidth]{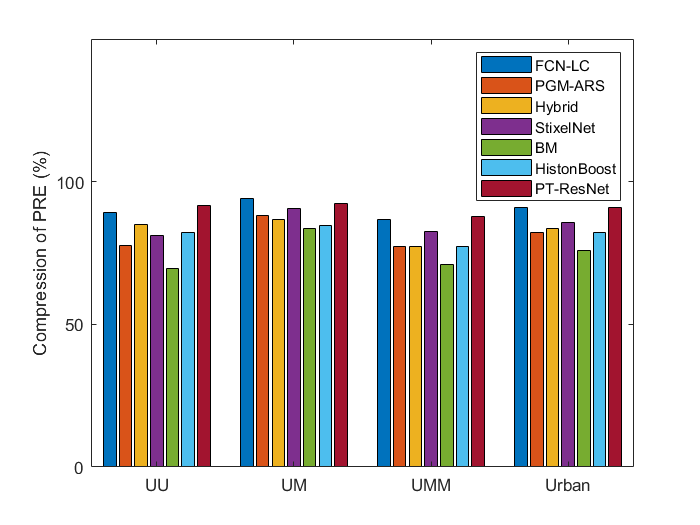}
		\label{fig.pre}}
	\subfigure[]
	{\includegraphics[width=0.35\textwidth]{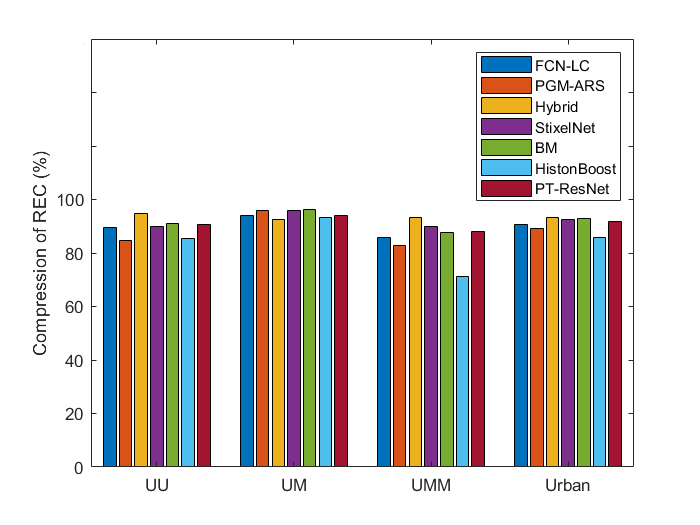}
		\label{fig.rec}}
	\\
	\subfigure[]
	{\includegraphics[width=0.35\textwidth]{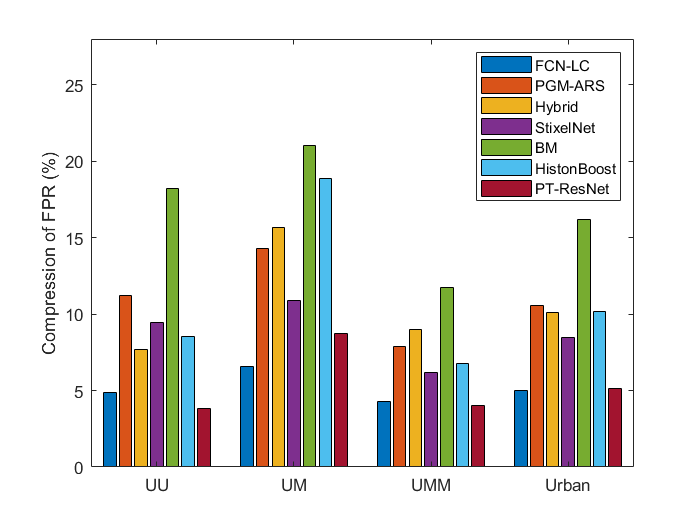}
		\label{fig.fpr}}
	\subfigure[]
	{\includegraphics[width=0.35\textwidth]{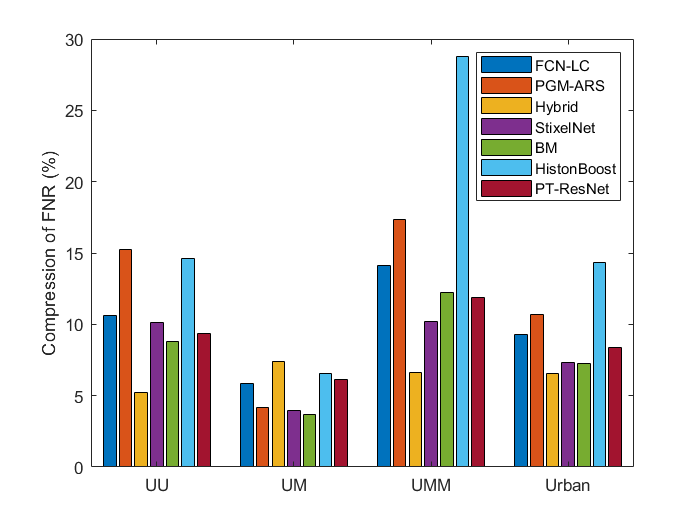}
		\label{fig.fnr}}
	\caption{Performance evaluation. (a) Comparison of MaxF. (b) Comparison of AP. (c) Comparison of PRE. (d) Comparison of REC. (e) Comparison of FPR. (f) Comparison of FNR.}
	\label{fig.comparison}
\end{figure*}

In recent years, the encoder-decoder structure has been prevalently used in deep neural networks for semantic segmentation \cite{deeplabv3+}. The encoder allows fast high-dimensional image feature map generation, while the decoder enables the network to recover sharp object boundaries \cite{deeplabv3+}. In this paper, our network is designed following ResNet-101 used in DeepLab-v3+ \cite{deeplabv3+}. The structure of our proposed  network is shown in Fig. \ref{fig.structure}.

\subsubsection{Encoder}
\label{sec.encoder}
In the encoder, the spatial dimension of the feature maps reduces gradually using four blocks, as shown in Fig. \ref{fig.structure}. The structure of each block is shown in Fig. \ref{fig.block_unit}, where BN denotes batch normalization, and ReLU denotes rectified linear unit. ReLU activation function is used to avoid overfitting during training. The parameter of ReLU is set to 0.5 in this paper. BN is a method used to normalize the input of each layer and overcome the internal covariate shift problem. As the stride in each block is set to 2, the output of the fourth block is 256 times smaller than the input of the first block. Furthermore, the baseline utilizes an atrous convolutional layer instead of a conventional convolutional layer. This allows us to enlarge the field-of-view of filters when interpolating the multi-scale context in the framework of spatial pyramid pooling and cascaded modules \cite{deeplabv3}. The output of each block can be computed by adding the output of the atrous convolutional layer to the input of the block, as shown in Fig. \ref{fig.block_unit}. 
The output of the fourth block feeds into five branches, as shown in Fig. \ref{fig.block_unit}. The baseline uses global average pooling to obtain the global image feature representations. In addition, three atrous convolutional layers with different rates are utilized to acquire multi-scale information. The rates depend entirely on the feature map produced by block 4, and they are set to 4, 8 and 16, respectively. Finally, the five branch output  is concatenated and further
 compressed using a $1\times1$ convolutional layer.  

\subsubsection{Decoder}
\label{sec.decoder}
In the decoder, the baseline applies skip connection to the feature map, which is produced by the second block. This greatly improves the details of local features in the high-level feature map. The low-level and high-level feature maps are then concatenated together. A probability map can be obtained after a feature map upsampling process. By finding the pixels whose probabilities are higher than our pre-set threshold, the semantic segmentation result can be obtained. Some examples of experimental results are shown in Fig. \ref{fig.experimental_results}.  

\section{Experimental Results}
\label{sec.experimental_results}
In this section, we present our experimental results and evaluate the performance of the proposed method using the KITTI  road dataset \cite{Fritsch2013}. The dataset contains synchronized stereo road image pairs, 3D road scenery point clouds acquired using a Velodyne HDL-64E LiDAR,  calibration parameters, and semantic region segmentation  ground truth. The images in this dataset are grouped into three categorizes: urban unmarked (UU), urban marked (UM) and urban multiple marked (UMM). To quantify the accuracy of the proposed approach, a set of indicators, including maximum F1-measure (MaxF), average precision (AP), precision (PRE), recall (REC), false positive rate (FPR) and false negative rate (FNR), are computed and are publicly available on the KITTI road benchmark\footnote{\url{http://www.cvlibs.net/datasets/kitti/eval_road.php.}}.   
PT-ResNet training was conducted on an NVIDIA GTX 1080 Ti GPU (CUDA 9 and cnDNN v7). In the experiments, the learning rate, training step and batch size are set to 0.001, 30000 and 8, respectively. The approach was programmed in Python language. The runtime of segmenting an image from the KITTI dataset  is around 3 seconds. 
In this section, we compare our method with FCN-LC \cite{Mendes2016}, PGM-ARS \cite{Passani2015}, Hybrid \cite{Xiao2018}, StixelNet \cite{Levi2015}, BM \cite{Wang2014} and HistonBoost \cite{Vitor2014}. The comparisons of MaxF, AP, PRE, REC, FPR and FNR among these methods are shown in Fig. \ref{fig.comparison}, where urban reflects the overall performance of UM, UMM and UU. It can be observed in Fig. \ref{fig.maxf} that  our PT-ResNet method outperforms the others in terms of MaxF, it achieves a MaxF of approximately $91.91\%$, which is slightly higher than that achieved using FCN-LC ($90.79\%$).  Fig. \ref{fig.ap} indicates that PT-ResNet performs better than other networks in terms of AP, as it achieves an AP of approximately $91.21\%$. However, FCN-LC performs slightly better than our network in terms of PRE and FPR (see Fig. \ref{fig.pre} and \ref{fig.fpr}). The overall PRE and FPR we achieved is $90.78\%$ and $5.13\%$, respectively.  Additionally, PT-ResNet achieves an intermediate performance in terms of REC and FNR (see Fig. \ref{fig.rec} and \ref{fig.fnr}), as the  REC and FNR values obtained using our method is $91.60\%$ and $8.40\%$, respectively.  In general, the proposed PT-ResNet achieves the best overall performance and its ranking is higher than that of other CNNs.
\section{Conclusion and Future Work}
\label{sec.conclusion}
In this paper, we presented a deep neural network for semantic road image segmentation. Since the proposed network focuses entirely on the road surface, the left and right stereo images were processed using our previously published perspective transformation algorithm. This greatly enhanced the similarity of the road surface between the left and right images. The processed stereo images and their corresponding subpixel disparity maps were utilized to create 3D training data. Additionally, we developed our network based on ResNet, a state-of-the-art network with  an encoder-decoder structure. According to the evaluation results provided by the KITTI road benchmark, our proposed method outperforms FCN-LC, PGM-ARS, Hybrid, StixelNet, BM and HistonBoost in terms of MaxF and AP, achieving an overall MaxF and AP of $91.19\%$ and $91.21\%$, respectively. 
However, the ResNet from DeepLab-v3+ may not be the best network for learning road semantic segmentation from our created 3D training data. Therefore, we plan to train different state-of-the-art networks, such as VGG-16 and VGG-19, and compare the experimental results with what we achieved in this paper. 

\section*{Acknowledgment}
This work was supported by the National Natural Science Foundation of China, under grant No. U1713211, the Research Grant Council of Hong Kong SAR Government, China, under Project No. 11210017, No. 21202816, and the Shenzhen Science, Technology and Innovation Commission (SZSTI) under grant JCYJ20160428154842603, awarded to Prof. Ming Liu.

\bibliographystyle{IEEEtran}
\balance

\end{document}